\documentclass[conference]{IEEEtran}
\IEEEoverridecommandlockouts
\usepackage{url}
\usepackage{cite}
\usepackage{amsmath,amssymb,amsfonts}
\usepackage{algorithmic}
\usepackage{graphicx}
\usepackage{textcomp}
\usepackage{xcolor}
\usepackage{graphicx}
\usepackage{caption}
\usepackage{lipsum}
\def\BibTeX{{\rm B\kern-.05em{\sc i\kern-.025em b}\kern-.08em
    T\kern-.1667em\lower.7ex\hbox{E}\kern-.125emX}}
\begin{document}

\title{AniArtAvatar: Animatable 3D Art Avatar from a Single Image
}

\author{\IEEEauthorblockN{Shaoxu Li}
\IEEEauthorblockA{\textit{John Hopcroft Center for Computer Science} \\
\textit{Shanghai Jiao Tong University}\\
Shanghai, China \\
lishaoxu@sjtu.edu.cn}

}

\twocolumn[{%
\renewcommand\twocolumn[1][]{#1}%
\maketitle
\begin{center}
    \centering
    \captionsetup{type=figure}
    \includegraphics[width=\textwidth]{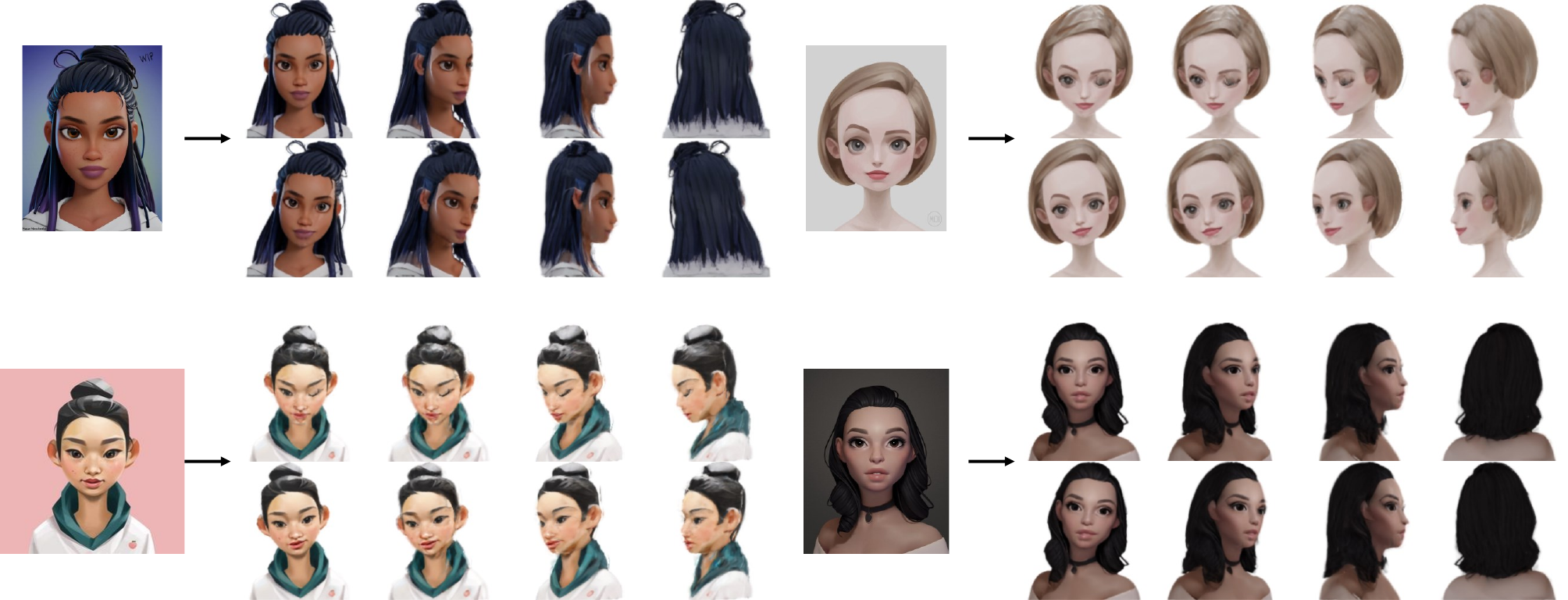}
    \captionof{figure}{In this work, we present AniArtAvatar, a novel one-shot 3D avatar generation and animation pipeline. The avatar can be animated with controllable 3D camera viewpoints, facial expressions, head poses, and shoulder movements. The method applies to various art styles.}
\end{center}%
}]


\begin{abstract}
We present a novel approach for generating animatable 3D-aware art avatars from a single image, with controllable facial expressions, head poses, and shoulder movements. Unlike previous reenactment methods, our approach utilizes a view-conditioned 2D diffusion model to synthesize multi-view images from a single art portrait with a neutral expression. With the generated colors and normals, we synthesize a static avatar using an SDF-based neural surface. For avatar animation, we extract control points, transfer the motion with these points, and deform the implicit canonical space. Firstly, we render the front image of the avatar, extract the 2D landmarks, and project them to the 3D space using a trained SDF network. We extract 3D driving landmarks using 3DMM and transfer the motion to the avatar landmarks. To animate the avatar pose, we manually set the body height and bound the head and torso of an avatar with two cages. The head and torso can be animated by transforming the two cages. Our approach is a one-shot pipeline that can be applied to various styles. Experiments demonstrate that our method can generate high-quality 3D art avatars with desired control over different motions.
\end{abstract}


\section{Introduction}
The digitization era has led to a surge in demand for automatic photo-realistic art avatar creation in various industries, such as computer gaming and film. However, creating 3D art characters is a complex and time-consuming task, even for experienced modeling artists. With the advancement of 3D scanning devices, generating realistic human digitization has become less challenging. A plethora of face and body datasets, such as 3DMM\cite{3dmm} and SMPL\cite{smpl}, are now available to support human digitization and animation. Additionally, utilizing neural implicit fields has enabled the achievement of photo-realistic rendering quality in avatar synthesis\cite{hong2021headnerf, Gafni_2021_Dynamic,grassal2021neural,zheng2022imavatar, ZielonkaCVPR2023INSTA}.

The progress in art avatar creation has yet to be entirely satisfactory, as the geometry may vary with styles. Art avatar creation can be divided into two tasks: stylization avatars, which have geometries similar to human models, and imaginary avatars, which may have different geometries from humans. Stylization avatar synthesis typically optimizes 3DMM or SMPL models with text or image inputs. For instance, AvatarCLIP\cite{hong2022avatarclip} initializes 3D avatars using the SMPL template mesh, optimizes a NeuS\cite{wang2021neus}, and styles the NeuS with CLIP and text prompts. An imaginary avatar synthesis pipeline typically involves training a model based on exaggerated 3D datasets. For example, StyleAvatar3D\cite{zhang2023styleavatar3d} utilizes image-text diffusion models to generate multi-view images of avatars and develops a latent diffusion model within the style space of StyleGAN\cite{Karras2019stylegan2} for avatar generation. Imaginary avatars are restricted to the domain of training data. While stylization avatars can synthesize arbitrary styles, they cannot achieve significant geometry warps.

The task of cross-domain face reenactment is related to our work, not considering 3D applications. This process involves transferring the motion of human portraits to art face images. To achieve this, a video dataset is required for motion pattern learning. Various methods, such as the transformer-based framework proposed by ToonTalker\cite{gong2023toontalker}, have been developed to align the motions from different domains in a shared canonical motion space for motion transfer. However, the typical face reenactment methods are limited to 2D and trained on cropped images, which restricts their application to wild images.

In this paper, we propose a novel one-shot method called AniArtAvatar for animatable avatar synthesis from a single art portrait. Our method allows for fine-grained control over facial expressions, head poses, and torso poses for various applications, and can be extended to exaggerated cartoon generation like local magnifications and illogical warps.

To achieve this, we utilize recent advances in zero-shot 3D generation with diffusion models. Specifically, we use conditioned diffusion models to obtain initial multi-view images from the input image. We then train an SDF-based implicit surface to synthesize the artistic avatar with these multi-view images. To animate the avatar, we render the front image, extract 2D landmarks, and project them to the implicit surface to obtain 3D landmarks. We then use the relative motion of human face landmarks to drive the 3D landmarks and deform the implicit surface for avatar expression animation. For avatar pose animation, we set the head and torso cages and execute transformations on them.

Our contributions are as follows:
\begin{itemize}
\item We propose a novel one-shot framework for animatable avatar synthesis. All we need is an input art portrait image.

\item We propose a pipeline for implicit surface-based art avatar animation in which expressions and poses can be controlled.

\item We conduct extensive experiments on various art styles to demonstrate the applicability and superiority of our method.
\end{itemize}

\section{Related Work}
\subsection{Zero-shot 3D Generation}
\subsubsection{3D Generation Guided by 2D Prior Models}
Generative models have been widely used to learn general representations of the world from large-scale 2D images. For instance, GAN-based methods, trained on specific domains, are capable of generating highly convincing images. Recently, diffusion and CLIP models have broken the domain boundaries of generative models. By training on Internet-scale image datasets, the diffusion model has made it possible to achieve universal image generation in any domain. Moreover, the powerful priors learned from wild images have made 3D generative tasks accessible. Several works, such as DreamField\cite{jain2021dreamfields}, DreamFusion\cite{poole2022dreamfusion}, Magic3D\cite{lin2023magic3d}, and subsequent text-to-3D works\cite{armandpour2023re,Chen_2023_Fantasia3D,chen2023it3d,huang2023dreamtime}, optimize a 3D representation with multi-view rendered images and text prompts. Another related task is image-to-3D, which aims to reconstruct a full 360° photographic model of an object from a single image. The pipeline for this task is similar to text-to-3D, and several works such as Realfusion\cite{melaskyriazi2023realfusion}, Magic123\cite{qian2023magic123}, Dreambooth3d\cite{raj2023dreambooth3d}, and others\cite{shen2023anything3d,tewari2023forwarddiffusion,tsalicoglou2023textmesh}, fit a 3D representation to the input image.

\subsubsection{Multi-view Diffusion Models}
Some methods explore multi-view image diffusion models to optimize the 3D scene indirectly. \cite{xiang2023ivid,zhang2023text2nerf} employ the diffusion model to infer the text-related image as content prior, use the depth estimation method to offer the geometric prior, and update the scene for novel view synthesis. Zero-1-to-3\cite{liu2023zero1to3} fine-tunes a pre-trained diffusion model on paired images, and their camera poses to learn controls over the camera parameters. However, Zero-1-to-3 cannot produce consistent results. One-2-3-45\cite{liu2023one2345} proposes a cost volume-based neural surface reconstruction module to handle inconsistent multi-view predictions. Viewset Diffusion\cite{szymanowicz23viewset_diffusion}, SyncDreamer\cite{liu2023syncdreamer}, MVDream\cite{shi2023MVDream} and Wonder3D\cite{long2023wonder3d} turn to attention layers to ensure multi-view consistency. Except for color images, Wonder3D also produces normals for explicit 3D extraction.

\subsection{Art Avatar Modeling and Generation}
As a hot topic, art avatar modeling and generation\cite{hong2022avatarclip,cao2023dreamavatar,kolotouros2023dreamhuman,shao2023control4d,li2023instructvideo2avatar,zhang2023styleavatar3d,luo2023rabit} has attracted much attention. We divide the generation of non-human avatars into stylization and imagination. Stylization modifies details on the human avatar while maintaining the overall geometry of the human avatar. Imaginary avatars have bold geometry that cannot be driven directly by human avatars.

\subsubsection{Stylization Avatar}
Some methods synthesize avatars from text prompts. AvatarCLIP\cite{hong2022avatarclip} and DreamAvatar\cite{cao2023dreamavatar} utilize CLIP to facilitate SMPL-based geometry sculpting and texture generation. Besides SMPL-based pose animation, DreamAvatar employs a landmark-based ControlNet\cite{zhang2023adding} to optimize the head pose and expression. AvatarVerse\cite{zhang2023avatarverse} takes text descriptions and pose guidance as a condition for avatar optimization. DreamHuman\cite{kolotouros2023dreamhuman} optimizes avatars with a deformable and pose-conditioned NeRF model learned and constrained using an implicit statistical 3D human pose and shape model. Rodin\cite{wang2022rodin} represents a neural radiance field as multiple 2D feature maps and rolls out these maps into a single 2D feature plane for perform 3D-aware diffusion.

Others modify a human avatar to stylized avatars. Control4D\cite{shao2023control4d} proposes to leverage a 4D generator to learn a more continuous generation space from inconsistent, edited images produced by the diffusion-based editor. Instruct-Video2Avatar\cite{li2023instructvideo2avatar} synthesizes an animatable avatar and then updates the rendered images with Instruct-Pix2Pix\cite{brooks2022instructpix2pix} for avatar stylization. Text2Control3D\cite{hwang2023text2control3d} extracts viewpoint-augmented depth maps and takes them as a condition of ControlNet for avatar synthesis. AlteredAvatar\cite{nguyenphuoc2023AA} uses the meta-learning framework to learn an initialization that can quickly adapt within a small number of update steps to a novel style, given a pre-captured avatar.

\subsubsection{Imaginary Avatar}
Imaginary character generation is a task similar to our method. However, few researchers tackle this task due to data limitations. To address this, 3DCaricShop\cite{qiu20213dcaricshop} introduces a 3D caricature dataset and presents a novel baseline approach for single-view 3D caricature reconstruction. SimpModeling\cite{luo2021simpmodeling} constructs a large manufactured anamorphic head dataset and presents a novel sketching system designed for amateur users to create desired anamorphic heads. StyleAvatar3D\cite{zhang2023styleavatar3d} employs poses extracted from existing 3D models to guide the generation of multi-view images and develops a coarse-to-fine discriminator for GAN training. RaBit\cite{luo2023rabit} proposes RaBit, the first 3D full-body cartoon parametric model for biped character modeling. Although these methods can facilitate the generation of imaginary avatars, they are restricted to the domain of the dataset. For arbitrary art character generation, these methods are powerless. In our work, we tackle this problem by utilizing the zero-shot image-to-3D diffusion-based method.

\begin{figure*}[t]
\centering
\includegraphics[width=1\textwidth]{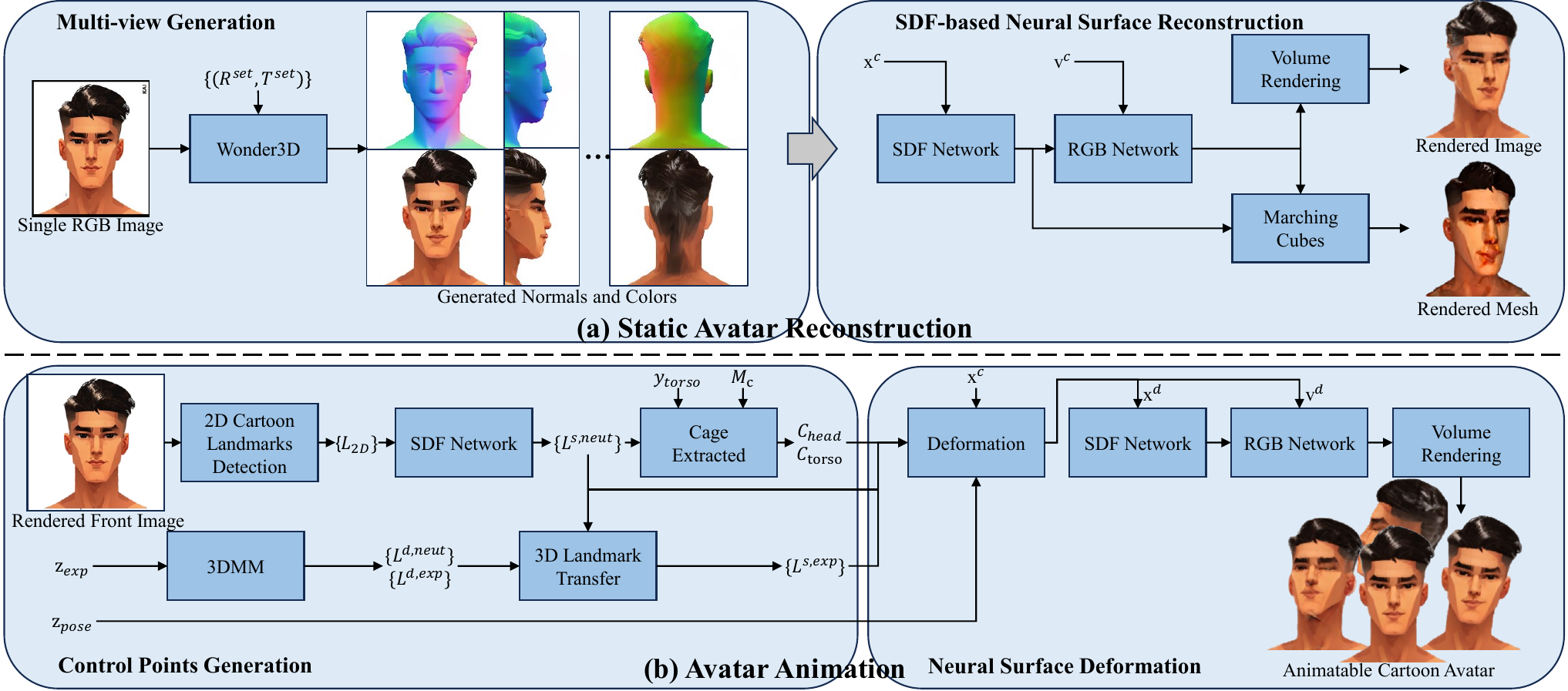} 
\caption{Method overview. (a) \textbf{Static Avatar Reconstruction}: Given an art portrait image, we use a view-conditioned 2D diffusion model, Wonder3D\cite{long2023wonder3d}, to generate multi-view images and normals. The input of Wonder3D includes a single image and pre-set camera poses. We optimize a static avatar using an SDF-based neural surface with generated images and normals. (b) \textbf{Avatar Animation}: We render the avatar into front image and extract 2D landmarks. With camera rays from 2D landmarks, we calculate the 3D corresponding landmarks $\{L^{s,neut}\}$ leveraging the SDF network. We transfer $\{L^{s,neut}\}$ to deformed landmarks $\{L^{s,exp}\}$ with drive landmarks $\{L^{d,neut}\}$ and $\{L^{d,exp}\}$. $\{L^{d,exp}\}$ can be manually set or extracted from 3DMM human face model. For head and shoulder movements, we extracted head cage $C_{head}$ and torso cage $C_{torso}$ using 3D landmarks $\{L^{s,neut}\}$, 3D canonical mesh $M_c$, and manually set torso height $y_{torso}$. We deform the canonical space for the static avatar to animate the art avatar with Delaunay triangulation-based deformation, with $\{L^{s,neut}\}$, $\{L^{s,exp}\}$, $C_{head}$, $C_{torso}$ and pose code $z_{pose}$ as input.} 
\label{fig:model}
\end{figure*}

\subsection{Face Reenactment}
\subsubsection{Within-domain Face Reenactment}
Face reenactment aims to produce natural talking head videos, given a source face image and driving talking head videos. Facial landmarks are a popular way to describe the geometry of human faces. Many methods\cite{wayne2018reenactgan,ZakharovFewShot,zhang2020freenet,ChenPuppeteerGAN} take the landmarks to guide the generation of reenactment videos. Compared with 2D landmarks, 3DMM models can capture intricate facial geometries well. A number of works\cite{ren2021pirenderer,Doukas_2021_ICCV,bounareli2022finding,YinStyleHEAT} animate the face with decoupled pose, expression, and identity with 3DMM. Unsupervised methods also play a role in face reenactment. Motion transfer assumes that the source and target objects have similar geometries. Based on this, many methods\cite{Siarohin_2019_fomm,SiarohinAnimating,wang2021oneshot,wiles2018x2face} learn the motion through unsupervised training.

\subsubsection{Cross-domain Face Reenactment}
The most similar task to our method, not considering the 3D requirements, is cross-domain face reenactment\cite{RecycleGAN,Yang2020MakeItTalk,Yang2020MakeItTalk,song2021pareidolia,xu2022motion,kim2021animeceleb,gong2023toontalker} that produces cartoon animation videos with driving human poses and expressions. Recycle-GAN\cite{RecycleGAN} incorporates spatiotemporal cues with conditional generative adversarial networks for video retargeting. New training is needed for a given new video pair. Make-it-talk\cite{Yang2020MakeItTalk} runs Delaunay triangulation on detected facial landmarks and guides the animation with the displaced landmarks. Similarly, Everythingtalking\cite{song2021pareidolia} proposes a parametric shape modeling on pareidolia face and transfer motions by facial landmarks. MAA\cite{xu2022motion} designs a shape-invariant motion adaptation module to capture the motion. AnimeCeleb\cite{kim2021animeceleb} utilizes 3D animation models as controllable image samplers to provide detailed pose annotations. ToonTalker\cite{gong2023toontalker} collects a cartoon video dataset in Disney style and proposes a transformer-based framework to align the motions from different domains. Although some progress has been achieved, arbitrary style artistic face reenactment is still a challenging task.

\section{Static Avatar Reconstruction}
We first present how our formulation learns the signed distance field of a static avatar from an art portrait. We synthesize multi-view avatar images and normals with a pre-trained conditioned diffusion model. Next, we synthesize an SDF-based neural surface with NeuS as the representation of an avatar.
\subsection{Wonder3D: View-Conditioned 2D Diffusion}
The recent work Wonder3D\cite{long2023wonder3d} proposes a multi-view cross-domain diffusion scheme that operates on two distinct domains to generate multi-view consistent normal maps and color images. Wonder3D proposes that the distribution of 3D assets, denoted as $p_a(z)$, can be modeled as a joint distribution of its corresponding 2D multi-view normal maps and corresponding color images:
\begin{equation}
    p_a(z)=p_{nc}(n^{1:K},i^{1:K}|i),
\end{equation}
where the distribution $p_{nc}$ of normal maps $n^{1:K}$ and color images $i^{1:K}$ are conditioned on an image $i$, given cameras $\{\pi_1,\pi_2\cdots\pi_K\}$. By learning the model 
\begin{equation}
    (n^{1:K},i^{1:K})=F(i,\pi_{1:K}),
\end{equation}
multi-view consistent normal maps and color images can be obtained from a single RGB image. The training is built upon the 2D diffusion models trained on billions of images like the Stable Diffusion model\cite{rombach2021highresolution,liu2022compositional,zhang2023adding}. For more details, we direct the reader to the original paper\cite{long2023wonder3d}. For static avatar reconstruction, we use the pre-trained Wonder3D to obtain the colors and normals for the following procedures, with a segmented art portrait as input.

\subsection{SDF-based Neural Surface Reconstruction}
Given calibrated multi-view images of a static scene, NeuS\cite{wang2021neus} implicitly represents the surface and appearance of a scene as a signed distance field(SDF) $S=\{f(\mathbf x)=0\}$ and a radiance field
\begin{equation}
    C(\mathbf o,\mathbf v)=\int_{0}^{\infty}w(t)c(p(t),\mathbf v)dt
\end{equation} 
implemented by MLPs, where $\mathbf o$ denotes view point, $\mathbf v$ denotes the view direction, $\mathbf x=p(t)=\mathbf o+\mathbf vt$ denotes 3D points along the ray, $c(p(t),\mathbf v)$ denotes the point color and $w(t)$ is a weighting function. Typically, SDF-based reconstruction methods require dense input views. With sparse input views, it is hard to reconstruct the scene well easily. Thanks to the normals from Wonder3D, a combination of RGB images and normals makes scene optimization possible.

After optimizing the avatar, we can obtain implicit avatars(neural surfaces) and explicit avatars(meshes). Although explicit meshes better fit existing workflows, the quality of extracted meshes may not be satisfactory. Considering this, we take the implicit avatars as our representation for animation, with the corresponding meshes as a supplement.

\section{Avatar Animation} For avatar facial animation, we assume the initial avatar has a neutral expression. We propose to warp the avatar through facial landmarks. For avatar pose animation, we propose to transform the head and torso and warp the neck for continuity. The warp is accomplished with Delaunay triangulation and deformation in 3D space.

\subsection{Avatar Expression Animation} To animate the avatar, we propose to deform the 3D space with some control points. For expression animation, facial landmarks are suitable. We use the popular 68 facial landmarks for avatar expression transfer.

\subsubsection{Landmarks Detection} To our knowledge, there is no previous work that detects 3D landmarks for arbitrary art portraits. To solve this, we propose to detect the 2D landmarks and project them into the 3D coordinates. For a static avatar, we render the front portrait image and 
train a 2D art face landmarks detection network\cite{Yaniv_2019_Face} to detect face landmarks for artistic images. For 3D landmarks $\{L^{s,neut}\}$ acquirement, rays from 2D landmarks $\{L_{2d}\}$ in the image are used to collide the SDF surface. For a 2D landmark pixel $o_L$ with direction $v$, the ray from it is denoted as $p(t)=\mathbf o+\mathbf vt$. The 3D landmark coordinate in the avatar scene is $\mathbf x_L=\arg\min\lvert f(\mathbf x) \rvert, \mathbf x\in p(t)$, where $f$ denotes the SDF network. For most facial areas, the pipeline works. However, landmarks outlining the whole chin may be outside the face area. The rest landmarks can be capable of representing the expressions well. Considering this, we only use 51 landmarks for expression animation, excluding 18 landmarks that outline the chin.

\begin{figure*}
    \includegraphics[width=1\textwidth]{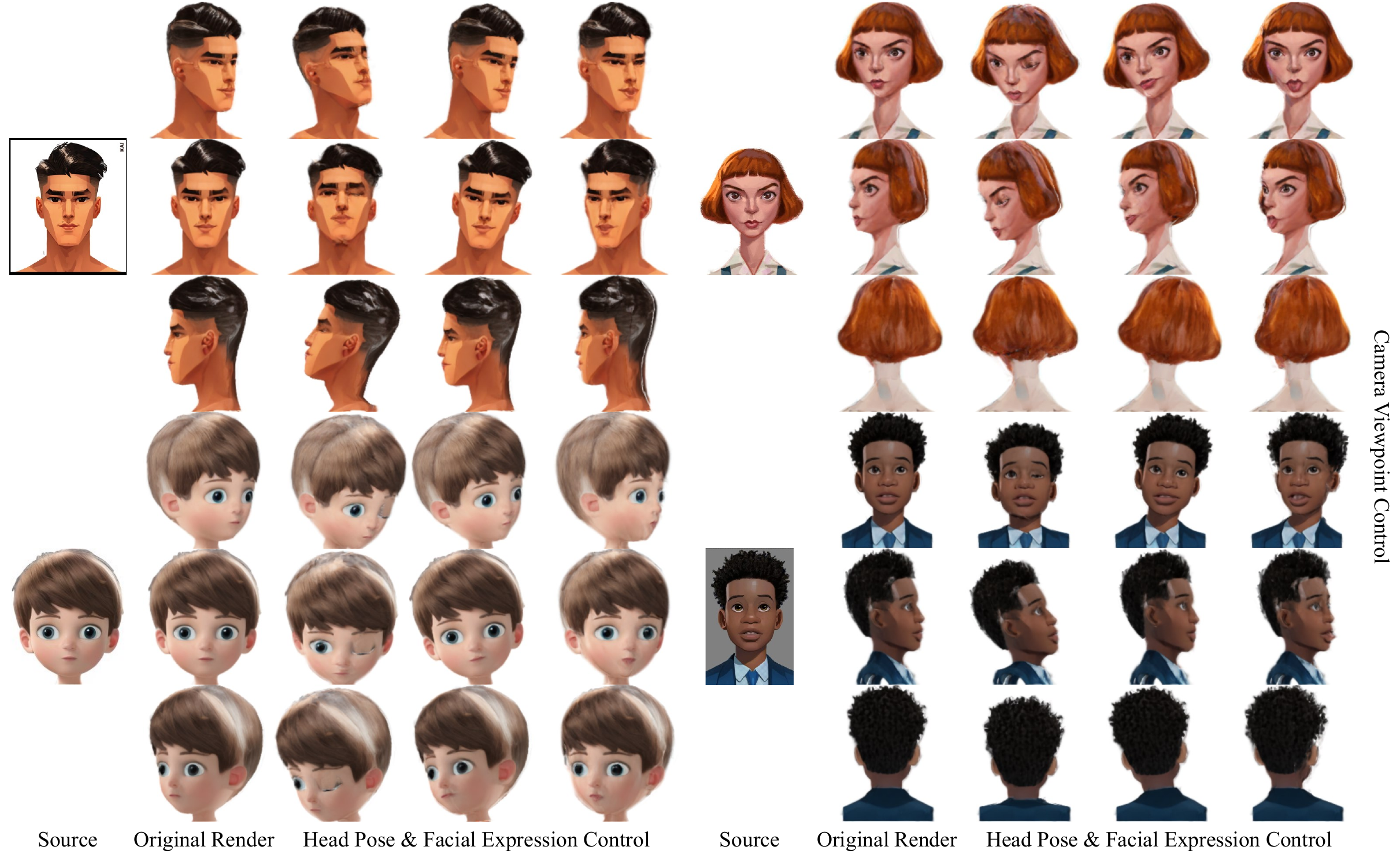} 
    \caption{Results with controllable camera viewpoint, facial expression and head pose.}
    \label{fig:exp1}
\end{figure*}

\subsubsection{Landmarks Deformation Transfer}
We assume the static avatar has a neutral expression. Given an expression code, we can obtain neutral and current 3D landmarks $\{L^{d,neut}\}$, $\{L^{d,exp}\}$ of the specific human faces. Now we transfer the neutral avatar landmarks $\{L^{s,neut}\}$ to expressive landmarks $\{L^{s,exp}\}$ to animate the avatar expression.

It is hard to learn the motion of arbitrary art portraits with a network. Inspired by\cite{Onizuka_2019_ICCV}, we directly transfer the 3D amount and direction of the movement. 
First, we organize the landmarks into six different parts $\{L\}_k,k\in[1:6]$. For expression transfer, we use 51 landmarks from the 68 landmarks(except the chin). The 51 landmarks include six facial parts: right eyebrow, left eyebrow, nose, right eye, left eye, and mouth. For art avatars, the size and orientation of the facial parts vary greatly. Considering this, we measure the size and orientation of each facial part and coherently adjust the amount and direction of the landmarks' motion. From each group, landmarks that can represent the orientation are selected, and we denote them as ${\{L'\}}_k$ with $N_k$ points. The orientation vector can be calculated by minimizing the following cost function, which ensures the normal vector $n_k$ becomes approximately perpendicular to all lines drawn between the landmarks in ${\{L'\}}_k$:
\begin{equation}
    E(n_k) = \sum^{N_k}_{i=0}\sum^{N_k}_{j=i}n_k(L'_i-L'_j),k\in[1:6]
\end{equation}
We compute the orientation vector $n_k$ of the six facial parts independently. With the six orientation vectors, we compute the rotation matrix $R_{k,s}$, $R_{k,d}$ that align the avatar and human face with the $z$ axis, respectively.
Then, we compute the 3D bounding box of the aligned facial parts with $a_{x,s},a_{y,s},a_{z,s}$ and $a_{x,d},a_{y,d},a_{z,d}$. The size ratios of avatar and human face are $A_{k,s}=diag(\frac{1}{a_{x,s}},\frac{1}{a_{y,s}},\frac{1}{a_{y,s}})$ and $A_{k,d}=diag(\frac{1}{a_{x,d}},\frac{a}{a_{y,d}},\frac{1}{a_{z,d}})$.

We then transfer the motion of landmarks by minimizing the following cost function:
\begin{equation}
E(\{L^{s,exp}\}) = \sum^6_{k=1}\sum_{i\in N_k}{A_{k,d}R_{k,d}u^d_i -A_{k,s}R_{k,s}u^s}
\label{eq:cost}
\end{equation}
where $u^d_i=L^{d,exp}_i -L^{d,neut}_i$, $u^s_i=L^{s,exp}_i -L^{s,neut}_i
$. We run Delaunay triangulation with the neutral 3D landmarks $\{L^{d,neut}\}$. By mapping the initial 3D space to the triangles, the subsequent animation process becomes straightforward. With deformed 3D landmarks $\{L^{s,exp}\}$, the deformed space represents the animated avatar.

\subsubsection{Additional Landmarks.} 
The SDF-based neural surface tries to reconstruct the scene with the surface. Unlike explicit meshes, the implicit surface cannot ensure that the density occupation is only on the surface. Therefore, when deforming the space, the avatar animation does not always work well. To ensure quality, a deformation of the space around the SDF surface is needed. In our task, we warp the space around the 3D facial landmarks in the same way instead of warping the surrounding space with triangular interpolation. To help the deformation of the scene, we propose adding additional landmarks. Specifically, we add additional key points perpendicular to the z-axis for each key point based on the camera pose of the front face.
\begin{equation}
    L_{add} = [L + [0,0,\alpha],L  - [0,0,\alpha]] 
\end{equation}
where $\alpha$ is an empirical value. In implementation, we add control points on the $\{L^{d,neut}\}$ and $\{L^{s,exp}\}$, respectively.

\begin{figure*}[t]
\centering
\includegraphics[width=1\textwidth]{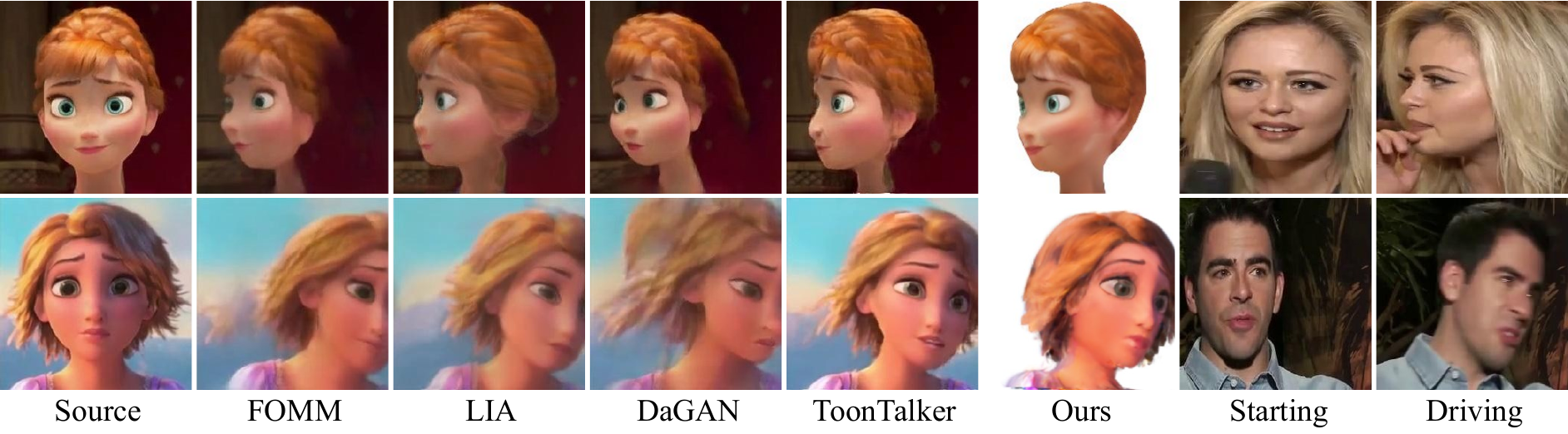} 
\caption{Qualitative comparisons with state-of-the-art methods on cross-domain reenactment. Competing methods are fine-tuned with the cartoon dataset.}
\label{fig:exp2}
\end{figure*}

\subsection{Avatar Pose Animation}
For avatar pose animation, we propose using two cages to guide the deformation of the implicit surface. Cage-based space deformation is a popular method for mesh and NeRF editing. Typically, NeRF editing extracts meshes, deforms the meshes, and propagates the deformation to the implicit space\cite{Yuan22NeRFEditing,xu2022deforming,peng2021CageNeRF}. The meshes are the cages for space deformation. Another way is to set some cages manually, which can maintain the geometry inside the cage stable\cite{li2023interactive}.

\subsubsection{Head Cage.}
We can easily obtain the head bottom position $y_{head}$ in the 2D front face, utilizing the art landmarks detection network. With the extracted mesh $M_c$ and head bottom position, the head cage is $C_{head}=\{\mathbf x\in M, \mathbf x[1]>y_{head}\}$.

\subsubsection{Torso Cage.}
There are many studies on the segment of the human body. However, the task is tough for art portraits with various geometries. Considering this, we propose that users can indicate the upper of the torso $y_{torso}$ in the rendered front image. It is a user-friendly interaction that only needs a click. With the extracted mesh $M_c$ and user-defined torso upper, the torso cage is 
$C_{torso}=\{\mathbf x\in M, \mathbf x[1]<y_{torso}\}$.

\subsubsection{Cages Transformation.}
Various operations are supported to the two cages, including translation, rotation, scaling, or any combinations. For points $\mathbf x$ in the cages, all the transformation takes the cage center as the central. The transformation can be denoted as $\mathbf x'=TRS\mathbf x_c$, where $\mathbf x'$ denotes the transformed coordinates, $\mathbf x_c$ denotes the center adjusted coordinates, $T$ denotes the translation matrix, $R$ denote the rotation matrices, $S$ denote the scale matrix.

For pose animation, the head and torso move. The neck shall be adjusted to ensure the continuity of the avatar. Similar to the expression animation, the space outside the head and torso cages is adjusted with Delaunay triangulation. We sample control points regularly on two cages and then warp the space between them. 

For realistic animation, we only use translation and rotation for cage transformation. For exaggerated animation, the scaling operation also plays a role.

\section{Experiments}
\textbf{Implementation Details}
Our method is a one-shot optimization-based method that does not need time-consuming training. All experiments are conducted on a single GeForce RTX3090 GPU. The initial outputs from Wonder3D\cite{long2023wonder3d} are images from six views with a resolution of 256×256, including the front, back, left, right, front-right, and front-left views. We train the SDF-based neural surface\cite{wang2021neus} with 10,000 steps and extract the mesh with 128×128 grids. We only implement the animation on the CPU, which takes about 5 minutes for one animation. The animation on GPU has yet to be completed.

\subsection{Generation Results}
Figure \ref{fig:teaser} and \ref{fig:exp1} present some generated art portraits from our method.  
For each case, the original render results show the multi-view render images of the static avatar. The other results show the animation results with different poses and expressions from the same view as the static avatar. The results show some representative expressions and poses. Expressions include left eye close, mouth left, and mouth open. Poses include head down and up, head left and right, and face left and right. Our method achieves consistent control for all animations for different art styles. More results can be found in the supplementary video.

\subsection{Comparison with Previous Methods}
We compare our method with four state-of-the-art face reenactment methods: FOMM\cite{Siarohin2019FOMM}, DAGAN\cite{hong2023dagan}, LIA\cite{wang2022latent} and ToonTalker\cite{gong2023toontalker}. To our knowledge, there is no previous work that deals with the animatable art avatar generation task in this paper. Hence, we compare ours with these four methods for reference purposes only. Our method applies to arbitrary art styles in a one-shot way. FOMM, DAGAN, LIA, and ToonTalker require training on datasets with specific art styles. For a fair comparison, we only compare results on cartoon datasets in Disney style\cite{gong2023toontalker}. Methods for comparison are trained or finetuned on the cartoon dataset.

\noindent\textbf{Qualitative Evaluation.}
Fig. \ref{fig:exp2} shows the reenactment results from real to cartoon.
Competing methods need start images and driving images. Our method only needs a single driving image, assuming the source image is with a neutral expression. Only our method produces 3D avatars.

\noindent\textbf{Quantitative Evaluation.}
Table \ref{tab:quantitative} shows the Frechet Inception Distance (FID)\cite{Heusel2017GANs} and Cumulative probability of Blur Detection (CPBD)\cite{Narvekar2011CPBD} metrics evaluated on these results. Our method achieves the best FID and CPBD, indicating that the synthetic results are most consistent with the source distribution with sharpness details. 

\begin{table}[htbp]
  \centering

    \scalebox{1.0}{\begin{tabular}{cccccc}
    & FOMM & LIA & DAGAN & ToonTalker & Ours\\
    \hline
    FID$\downarrow$ &  40.853 & 34.066 & 36.218 & 27.467 & 23.756 \\ 
    CPBD$\uparrow$ &   0.0368 & 0.0523 & 0.0598 & 0.0846 & 0.1207 \\ 
    \hline
    \end{tabular}%
    }
    \caption{Quantitative comparisons on cross-domain reenactment. Competing methods are fine-tuned with the cartoon dataset.}
  \label{tab:quantitative}%
\end{table}%


\subsection{Ablation Study}
We conduct ablation studies to validate the effectiveness of our algorithm design.

\begin{figure}[h]
  \centering
  \includegraphics[width=0.92\linewidth]{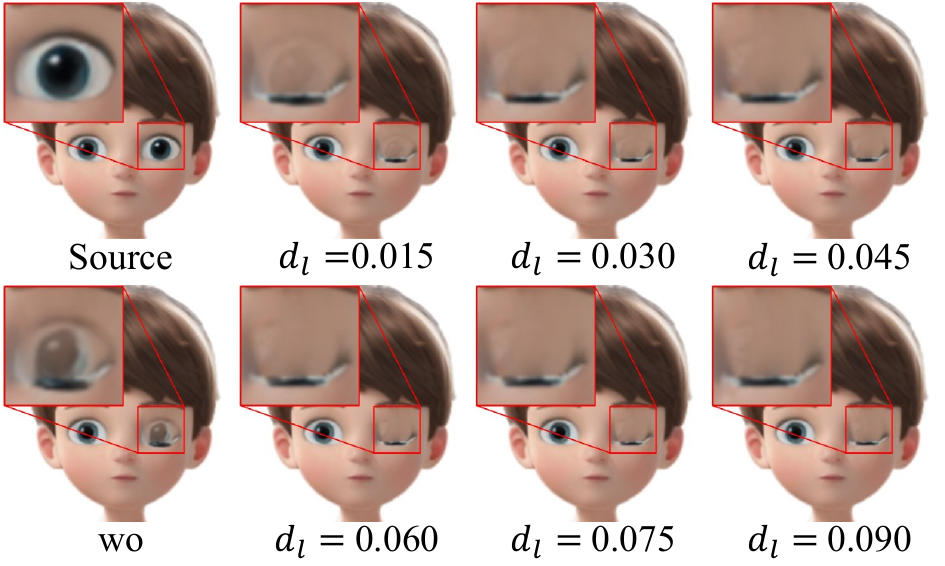}
  \caption{Ablation study on additional landmarks.}
  \label{fig:ablation_lms}
\end{figure}

\noindent\textbf{Additional Landmarks.}
We propose adding additional landmarks to improve the visual quality of expression animation. Figure \ref{fig:ablation_lms} shows the results. The visual quality improves with the distance $d_l$ of additional landmarks. We chose $d_l=0.45$ to find the balance between deformation and the influence areas. From 0.45 to 0.90, the results change slightly. Therefore, we chose 0.45 as our additional landmarks distance.

\begin{figure}[h]
  \centering
  \includegraphics[width=\linewidth]{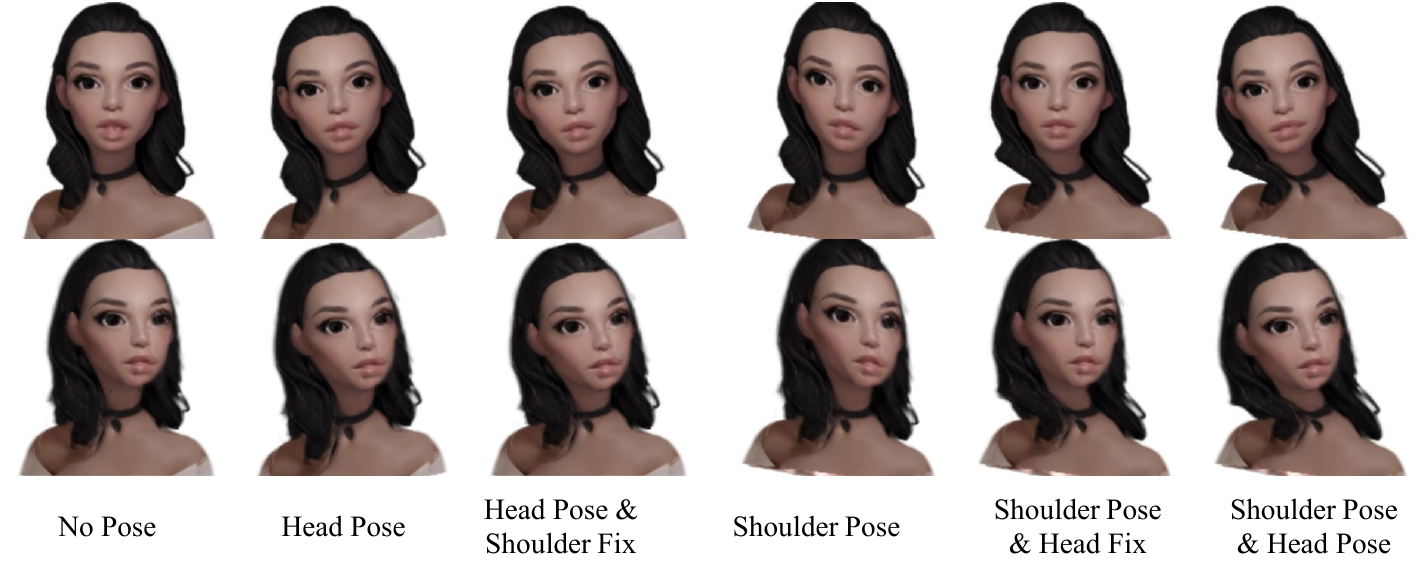}
  \caption{Ablation study on head cage and torso cage.}
  \label{fig:ablation_cage}
\end{figure}

\noindent\textbf{Head Cage and Torso Cage.}
Our method reconstructs the avatar with a torso cage and a head cage. However, our method cannot handle areas beyond the source image, and torso animation cannot always promise the visual quality of the avatar. Fig. \ref{fig:ablation_cage} shows the results. We can animate the head and torso separately or together. When animating one, the other should be fixed to ensure the ideal animation. We recommend animating the head and keeping the shoulder fixed.

\begin{figure}[h]
  \centering
  \includegraphics[width=\linewidth]{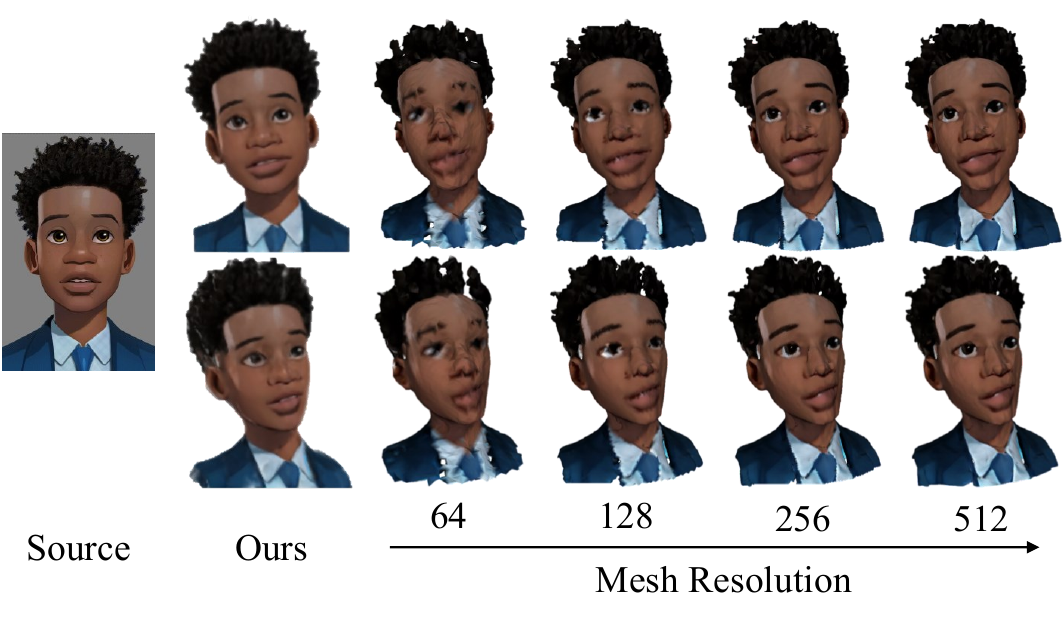}
  \caption{Ablation study on neural surface representation.}
  \label{fig:ablation_mesh}
\end{figure}

\noindent\textbf{Neural Surface Representation.}
Although implicit representation has attracted much attention for its high quality, explicit models are still the dominant approach in engineering. It is natural to extract meshes of the avatar and deform the meshes to animate the avatar. However, in our task, the mesh quality is far from the application. Fig. \ref{fig:ablation_mesh} shows the results. With the increase in mesh resolution (resolution of marching cubes), the quality of rendered meshes improved. For the maximum 512 resolution, which has more than 460,000 vertices, the rendering quality is still far from matching the implicit surface.

\subsection{Discussion and Limitations}
\noindent\textbf{Discussion.} For avatar animation, we transfer the relative motion of human faces to art avatars. For human faces, the neutral expression can be obtained from 3DMM face models. However, for art faces, it is not easy to estimate the neutral expression from expressive faces. Therefore, our method assumes that the source portrait has a neutral expression and pose. We use an implicit surface to synthesize the art avatar. Although high-quality results can be obtained, the result is less compatible with existing workflows. The easy application of implicit avatars remains to be studied.

\begin{figure}[h]
  \centering
  \includegraphics[width=\linewidth]{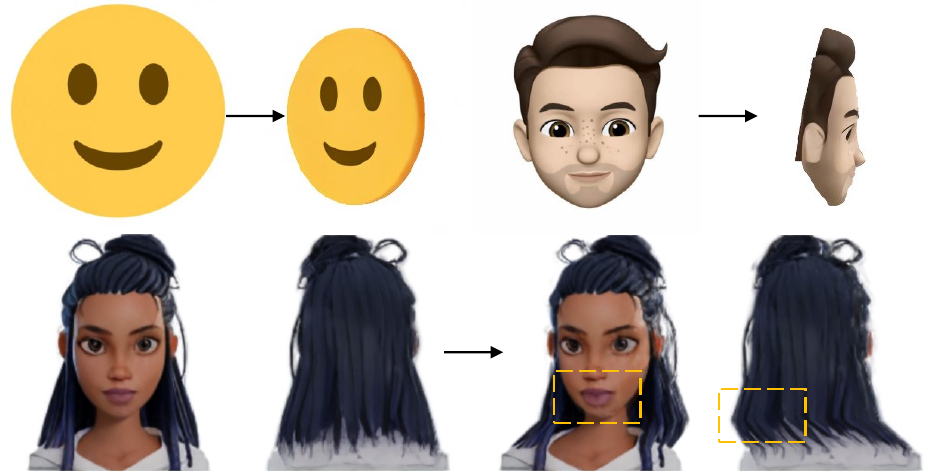}
  \caption{Limitations on some 2D images, unsatisfactory open mouth, and hair warps. The first line shows two cases from source images to 3D results. The second line shows cases from original render results to animated avatars.}
  \label{fig:limitations}
\end{figure}

\noindent\textbf{Limitations.} Due to the limited ability of zero-shot 3D generation methods, the results are not perfect for some 2D art images. Although results can be obtained, the 3D geometries are not as expected, as shown in Fig. \ref{fig:limitations}. Additionally, our method synthesizes the avatar as a normal scene, which cannot reconstruct the inner geometry of the face. The animation only deforms the avatar without creation. Due to the lack of geometry and texture, the visual quality of the inner mouth region (e.g., teeth) is not satisfactory. Moreover, the motion patterns of some cartoon animations are different from those of humans. For example, the mouth may be unnaturally large when laughing, which is beyond the scope of our method.

\section{Conclusion}
In this paper, we present a novel one-shot pipeline, AniArtAvatar, for generating animatable head-shoulder 3D art avatars. To generate the avatars, we propose synthesizing an SDF-based implicit surface by leveraging pre-trained diffusion-based multi-view image synthesis models. We extract 3D landmarks by detecting 2D landmarks and projecting them onto the implicit surface. Expression animation is achieved through landmark transfer and Delaunay triangulation-based deformation, while pose animation is accomplished through head and torso cage transformation and neck deformation. We conducted extensive experiments to demonstrate the superiority of our method.
\bibliographystyle{IEEEbib}
\bibliography{icme2023template}

\begin{thebibliography}{10}

\bibitem{3dmm}
Volker Blanz and Thomas Vetter,
\newblock ``A morphable model for the synthesis of 3d faces,''
\newblock in {\em Proceedings of the 26th Annual Conference on Computer Graphics and Interactive Techniques}, USA, 1999, SIGGRAPH '99, p. 187–194, ACM Press/Addison-Wesley Publishing Co.

\bibitem{smpl}
Matthew Loper, Naureen Mahmood, Javier Romero, Gerard Pons-Moll, and Michael~J. Black,
\newblock ``Smpl: A skinned multi-person linear model,''
\newblock {\em ACM Trans. Graph.}, vol. 34, no. 6, oct 2015.

\bibitem{hong2021headnerf}
Yang Hong, Bo~Peng, Haiyao Xiao, Ligang Liu, and Juyong Zhang,
\newblock ``Headnerf: A real-time nerf-based parametric head model,''
\newblock in {\em {IEEE/CVF} Conference on Computer Vision and Pattern Recognition (CVPR)}, 2022.

\bibitem{Gafni_2021_Dynamic}
Guy Gafni, Justus Thies, Michael Zollh{\"o}fer, and Matthias Nie{\ss}ner,
\newblock ``Dynamic neural radiance fields for monocular 4d facial avatar reconstruction,''
\newblock in {\em Proceedings of the IEEE/CVF Conference on Computer Vision and Pattern Recognition (CVPR)}, June 2021, pp. 8649--8658.

\bibitem{grassal2021neural}
Philip-William Grassal, Malte Prinzler, Titus Leistner, Carsten Rother, Matthias Nie{\ss}ner, and Justus Thies,
\newblock ``Neural head avatars from monocular rgb videos,''
\newblock {\em arXiv preprint arXiv:2112.01554}, 2021.

\bibitem{zheng2022imavatar}
Yufeng Zheng, Victoria~Fernández Abrevaya, Marcel~C. Bühler, Xu~Chen, Michael~J. Black, and Otmar Hilliges,
\newblock ``{I} {M} {Avatar}: Implicit morphable head avatars from videos,''
\newblock in {\em Computer Vision and Pattern Recognition (CVPR)}, 2022.

\bibitem{ZielonkaCVPR2023INSTA}
Wojciech Zielonka, Timo Bolkart, and Justus Thies,
\newblock ``Instant volumetric head avatars,''
\newblock in {\em CVPR}, 2023.

\bibitem{hong2022avatarclip}
Fangzhou Hong, Mingyuan Zhang, Liang Pan, Zhongang Cai, Lei Yang, and Ziwei Liu,
\newblock ``Avatarclip: Zero-shot text-driven generation and animation of 3d avatars,''
\newblock {\em ACM Transactions on Graphics (TOG)}, vol. 41, no. 4, pp. 1--19, 2022.

\bibitem{wang2021neus}
Peng Wang, Lingjie Liu, Yuan Liu, Christian Theobalt, Taku Komura, and Wenping Wang,
\newblock ``Neus: Learning neural implicit surfaces by volume rendering for multi-view reconstruction,''
\newblock {\em arXiv preprint arXiv:2106.10689}, 2021.

\bibitem{zhang2023styleavatar3d}
Chi Zhang, Yiwen Chen, Yijun Fu, Zhenglin Zhou, Gang YU, Billzb Wang, Bin Fu, Tao Chen, Guosheng Lin, and Chunhua Shen,
\newblock ``Styleavatar3d: Leveraging image-text diffusion models for high-fidelity 3d avatar generation,'' 2023.

\bibitem{Karras2019stylegan2}
Tero Karras, Samuli Laine, Miika Aittala, Janne Hellsten, Jaakko Lehtinen, and Timo Aila,
\newblock ``Analyzing and improving the image quality of {StyleGAN},''
\newblock in {\em Proc. CVPR}, 2020.

\bibitem{gong2023toontalker}
Gong Yuan, Zhang Yong, Cun Xiaodong, Yin Fei, Fan Yanbo, Wang Xuan, Wu~Baoyuan, and Yang Yujiu,
\newblock ``Toontalker: Cross-domain face reenactment,'' 2023.

\bibitem{jain2021dreamfields}
Ajay Jain, Ben Mildenhall, Jonathan~T. Barron, Pieter Abbeel, and Ben Poole,
\newblock ``Zero-shot text-guided object generation with dream fields,''
\newblock 2022.

\bibitem{poole2022dreamfusion}
Ben Poole, Ajay Jain, Jonathan~T Barron, and Ben Mildenhall,
\newblock ``Dreamfusion: Text-to-3d using 2d diffusion,''
\newblock {\em arXiv preprint arXiv:2209.14988}, 2022.

\bibitem{lin2023magic3d}
Chen-Hsuan Lin, Jun Gao, Luming Tang, Towaki Takikawa, Xiaohui Zeng, Xun Huang, Karsten Kreis, Sanja Fidler, Ming-Yu Liu, and Tsung-Yi Lin,
\newblock ``Magic3d: High-resolution text-to-3d content creation,''
\newblock in {\em Proceedings of the IEEE/CVF Conference on Computer Vision and Pattern Recognition}, 2023, pp. 300--309.

\bibitem{armandpour2023re}
Mohammadreza Armandpour, Huangjie Zheng, Ali Sadeghian, Amir Sadeghian, and Mingyuan Zhou,
\newblock ``Re-imagine the negative prompt algorithm: Transform 2d diffusion into 3d, alleviate janus problem and beyond,''
\newblock {\em arXiv preprint arXiv:2304.04968}, 2023.

\bibitem{Chen_2023_Fantasia3D}
Rui Chen, Yongwei Chen, Ningxin Jiao, and Kui Jia,
\newblock ``Fantasia3d: Disentangling geometry and appearance for high-quality text-to-3d content creation,''
\newblock in {\em Proceedings of the IEEE/CVF International Conference on Computer Vision (ICCV)}, October 2023.

\bibitem{chen2023it3d}
Yiwen Chen, Chi Zhang, Xiaofeng Yang, Zhongang Cai, Gang Yu, Lei Yang, and Guosheng Lin,
\newblock ``It3d: Improved text-to-3d generation with explicit view synthesis,'' 2023.

\bibitem{huang2023dreamtime}
Yukun Huang, Jianan Wang, Yukai Shi, Xianbiao Qi, Zheng-Jun Zha, and Lei Zhang,
\newblock ``Dreamtime: An improved optimization strategy for text-to-3d content creation,'' 2023.

\bibitem{melaskyriazi2023realfusion}
Luke Melas-Kyriazi, Christian Rupprecht, Iro Laina, and Andrea Vedaldi,
\newblock ``Realfusion: 360° reconstruction of any object from a single image,''
\newblock in {\em Arxiv}, 2023.

\bibitem{qian2023magic123}
Guocheng Qian, Jinjie Mai, Abdullah Hamdi, Jian Ren, Aliaksandr Siarohin, Bing Li, Hsin-Ying Lee, Ivan Skorokhodov, Peter Wonka, Sergey Tulyakov, and Bernard Ghanem,
\newblock ``Magic123: One image to high-quality 3d object generation using both 2d and 3d diffusion priors,''
\newblock {\em arXiv preprint arXiv:2306.17843}, 2023.

\bibitem{raj2023dreambooth3d}
Amit Raj, Srinivas Kaza, Ben Poole, Michael Niemeyer, Ben Mildenhall, Nataniel Ruiz, Shiran Zada, Kfir Aberman, Michael Rubenstein, Jonathan Barron, Yuanzhen Li, and Varun Jampani,
\newblock ``Dreambooth3d: Subject-driven text-to-3d generation,''
\newblock {\em ICCV}, 2023.

\bibitem{shen2023anything3d}
Qiuhong Shen, Xingyi Yang, and Xinchao Wang,
\newblock ``Anything-3d: Towards single-view anything reconstruction in the wild,'' 2023.

\bibitem{tewari2023forwarddiffusion}
Ayush Tewari, Tianwei Yin, George Cazenavette, Semon Rezchikov, Joshua~B. Tenenbaum, Frédo Durand, William~T. Freeman, and Vincent Sitzmann,
\newblock ``Diffusion with forward models: Solving stochastic inverse problems without direct supervision,''
\newblock in {\em arXiv}, 2023.

\bibitem{tsalicoglou2023textmesh}
Christina Tsalicoglou, Fabian Manhardt, Alessio Tonioni, Michael Niemeyer, and Federico Tombari,
\newblock ``Textmesh: Generation of realistic 3d meshes from text prompts,''
\newblock {\em arXiv preprint arXiv:2304.12439}, 2023.

\bibitem{xiang2023ivid}
Jianfeng Xiang, Jiaolong Yang, Binbin Huang, and Xin Tong,
\newblock ``3d-aware image generation using 2d diffusion models,''
\newblock in {\em Proceedings of the IEEE/CVF International Conference on Computer Vision (ICCV)}, October 2023, pp. 2383--2393.

\bibitem{zhang2023text2nerf}
Jingbo Zhang, Xiaoyu Li, Ziyu Wan, Can Wang, and Jing Liao,
\newblock ``Text2nerf: Text-driven 3d scene generation with neural radiance fields,'' 2023.

\bibitem{liu2023zero1to3}
Ruoshi Liu, Rundi Wu, Basile~Van Hoorick, Pavel Tokmakov, Sergey Zakharov, and Carl Vondrick,
\newblock ``Zero-1-to-3: Zero-shot one image to 3d object,'' 2023.

\bibitem{liu2023one2345}
Minghua Liu, Chao Xu, Haian Jin, Linghao Chen, Zexiang Xu, Hao Su, et~al.,
\newblock ``One-2-3-45: Any single image to 3d mesh in 45 seconds without per-shape optimization,''
\newblock {\em arXiv preprint arXiv:2306.16928}, 2023.

\bibitem{szymanowicz23viewset_diffusion}
Stanislaw Szymanowicz, Christian Rupprecht, and Andrea Vedaldi,
\newblock ``Viewset diffusion: (0-)image-conditioned {3D} generative models from {2D} data,''
\newblock in {\em ICCV}, 2023.

\bibitem{liu2023syncdreamer}
Yuan Liu, Cheng Lin, Zijiao Zeng, Xiaoxiao Long, Lingjie Liu, Taku Komura, and Wenping Wang,
\newblock ``Syncdreamer: Generating multiview-consistent images from a single-view image,''
\newblock {\em arXiv preprint arXiv:2309.03453}, 2023.

\bibitem{shi2023MVDream}
Yichun Shi, Peng Wang, Jianglong Ye, Long Mai, Kejie Li, and Xiao Yang,
\newblock ``Mvdream: Multi-view diffusion for 3d generation,''
\newblock {\em arXiv:2308.16512}, 2023.

\bibitem{long2023wonder3d}
Xiaoxiao Long, Yuan-Chen Guo, Cheng Lin, Yuan Liu, Zhiyang Dou, Lingjie Liu, Yuexin Ma, Song-Hai Zhang, Marc Habermann, Christian Theobalt, et~al.,
\newblock ``Wonder3d: Single image to 3d using cross-domain diffusion,''
\newblock {\em arXiv preprint arXiv:2310.15008}, 2023.

\bibitem{cao2023dreamavatar}
Yukang Cao, Yan-Pei Cao, Kai Han, Ying Shan, and Kwan-Yee~K Wong,
\newblock ``Dreamavatar: Text-and-shape guided 3d human avatar generation via diffusion models,''
\newblock {\em arXiv preprint arXiv:2304.00916}, 2023.

\bibitem{kolotouros2023dreamhuman}
Nikos Kolotouros, Thiemo Alldieck, Andrei Zanfir, Eduard~Gabriel Bazavan, Mihai Fieraru, and Cristian Sminchisescu,
\newblock ``Dreamhuman: Animatable 3d avatars from text,''
\newblock 2023.

\bibitem{shao2023control4d}
Ruizhi Shao, Jingxiang Sun, Cheng Peng, Zerong Zheng, Boyao Zhou, Hongwen Zhang, and Yebin Liu,
\newblock ``Control4d: Efficient 4d portrait editing with text,''
\newblock 2023.

\bibitem{li2023instructvideo2avatar}
Shaoxu Li,
\newblock ``Instruct-video2avatar: Video-to-avatar generation with instructions,'' 2023.

\bibitem{luo2023rabit}
Zhongjin Luo, Shengcai Cai, Jinguo Dong, Ruibo Ming, Liangdong Qiu, Xiaohang Zhan, and Xiaoguang Han,
\newblock ``Rabit: Parametric modeling of 3d biped cartoon characters with a topological-consistent dataset,''
\newblock in {\em Proceedings of the IEEE/CVF Conference on Computer Vision and Pattern Recognition (CVPR)}, 2023.

\bibitem{zhang2023adding}
Lvmin Zhang, Anyi Rao, and Maneesh Agrawala,
\newblock ``Adding conditional control to text-to-image diffusion models,'' 2023.

\bibitem{zhang2023avatarverse}
Huichao Zhang, Bowen Chen, Hao Yang, Liao Qu, Xu~Wang, Li~Chen, Chao Long, Feida Zhu, Kang Du, and Min Zheng,
\newblock ``Avatarverse: High-quality \& stable 3d avatar creation from text and pose,'' 2023.

\bibitem{wang2022rodin}
Tengfei Wang, Bo~Zhang, Ting Zhang, Shuyang Gu, Jianmin Bao, Tadas Baltrusaitis, Jingjing Shen, Dong Chen, Fang Wen, Qifeng Chen, and Baining Guo,
\newblock ``Rodin: A generative model for sculpting 3d digital avatars using diffusion,'' 2022.

\bibitem{brooks2022instructpix2pix}
Tim Brooks, Aleksander Holynski, and Alexei~A. Efros,
\newblock ``Instructpix2pix: Learning to follow image editing instructions,''
\newblock in {\em CVPR}, 2023.

\bibitem{hwang2023text2control3d}
Sungwon Hwang, Junha Hyung, and Jaegul Choo,
\newblock ``Text2control3d: Controllable 3d avatar generation in neural radiance fields using geometry-guided text-to-image diffusion model,'' 2023.

\bibitem{nguyenphuoc2023AA}
Thu Nguyen-Phuoc, Gabriel Schwartz, Yuting Ye, Stephen Lombardi, and Lei Xiao,
\newblock ``Alteredavatar: Stylizing dynamic 3d avatars with fast style adaptation,''
\newblock {\em arXiv:2305.19245}, 2023.

\bibitem{qiu20213dcaricshop}
Yuda Qiu, Xiaojie Xu, Lingteng Qiu, Yan Pan, Yushuang Wu, Weikai Chen, and Xiaoguang Han,
\newblock ``3dcaricshop: A dataset and a baseline method for single-view 3d caricature face reconstruction,''
\newblock in {\em Proceedings of the IEEE/CVF Conference on Computer Vision and Pattern Recognition}, 2021, pp. 10236--10245.

\bibitem{luo2021simpmodeling}
Zhongjin Luo, Jie Zhou, Heming Zhu, Dong Du, Xiaoguang Han, and Hongbo Fu,
\newblock ``Simpmodeling: Sketching implicit field to guide mesh modeling for 3d animalmorphic head design,''
\newblock in {\em The 34th Annual ACM Symposium on User Interface Software and Technology}, 2021, pp. 854--863.

\bibitem{wayne2018reenactgan}
Wayne Wu, Yunxuan Zhang, Cheng Li, Chen Qian, and Chen~Change Loy,
\newblock ``Reenactgan: Learning to reenact faces via boundary transfer,''
\newblock in {\em ECCV}, September 2018.

\bibitem{ZakharovFewShot}
Egor Zakharov, Aliaksandra Shysheya, Egor Burkov, and Victor Lempitsky,
\newblock ``Few-shot adversarial learning of realistic neural talking head models,''
\newblock in {\em 2019 IEEE/CVF International Conference on Computer Vision (ICCV)}, 2019, pp. 9458--9467.

\bibitem{zhang2020freenet}
Jiangning Zhang, Xianfang Zeng, Mengmeng Wang, Yusu Pan, Liang Liu, Yong Liu, Yu~Ding, and Changjie Fan,
\newblock ``Freenet: Multi-identity face reenactment,''
\newblock in {\em CVPR}, 2020, pp. 5326--5335.

\bibitem{ChenPuppeteerGAN}
Zhuo Chen, Chaoyue Wang, Bo~Yuan, and Dacheng Tao,
\newblock ``Puppeteergan: Arbitrary portrait animation with semantic-aware appearance transformation,''
\newblock in {\em 2020 IEEE/CVF Conference on Computer Vision and Pattern Recognition (CVPR)}, 2020, pp. 13515--13524.

\bibitem{ren2021pirenderer}
Yurui Ren, Ge~Li, Yuanqi Chen, Thomas~H. Li, and Shan Liu,
\newblock ``Pirenderer: Controllable portrait image generation via semantic neural rendering,'' 2021.

\bibitem{Doukas_2021_ICCV}
Michail~Christos Doukas, Stefanos Zafeiriou, and Viktoriia Sharmanska,
\newblock ``Headgan: One-shot neural head synthesis and editing,''
\newblock in {\em Proceedings of the IEEE/CVF International Conference on Computer Vision (ICCV)}, October 2021, pp. 14398--14407.

\bibitem{bounareli2022finding}
Stella Bounareli, Vasileios Argyriou, and Georgios Tzimiropoulos,
\newblock ``Finding directions in gan's latent space for neural face reenactment,'' 2022.

\bibitem{YinStyleHEAT}
Fei Yin, Yong Zhang, Xiaodong Cun, Mingdeng Cao, Yanbo Fan, Xuan Wang, Qingyan Bai, Baoyuan Wu, Jue Wang, and Yujiu Yang,
\newblock ``Styleheat: One-shot high-resolution editable talking face generation via pre-trained stylegan,''
\newblock {\em arxiv:2203.04036}, 2022.

\bibitem{Siarohin_2019_fomm}
Aliaksandr Siarohin, Stéphane Lathuilière, Sergey Tulyakov, Elisa Ricci, and Nicu Sebe,
\newblock ``First order motion model for image animation,''
\newblock in {\em Conference on Neural Information Processing Systems (NeurIPS)}, December 2019.

\bibitem{SiarohinAnimating}
Aliaksandr Siarohin, Stéphane Lathuilière, Sergey Tulyakov, Elisa Ricci, and Nicu Sebe,
\newblock ``Animating arbitrary objects via deep motion transfer,''
\newblock in {\em 2019 IEEE/CVF Conference on Computer Vision and Pattern Recognition (CVPR)}, 2019, pp. 2372--2381.

\bibitem{wang2021oneshot}
Ting-Chun Wang, Arun Mallya, and Ming-Yu Liu,
\newblock ``One-shot free-view neural talking-head synthesis for video conferencing,'' 2021.

\bibitem{wiles2018x2face}
Olivia Wiles, A.~Sophia Koepke, and Andrew Zisserman,
\newblock ``X2face: A network for controlling face generation by using images, audio, and pose codes,'' 2018.

\bibitem{RecycleGAN}
Aayush Bansal, Shugao Ma, Deva Ramanan, and Yaser Sheikh,
\newblock ``Recycle-gan: Unsupervised video retargeting,''
\newblock in {\em ECCV}, 2018.

\bibitem{Yang2020MakeItTalk}
Yang Zhou, Xintong Han, Eli Shechtman, Jose Echevarria, Evangelos Kalogerakis, and Dingzeyu Li,
\newblock ``Makeittalk: Speaker-aware talking-head animation,''
\newblock {\em ACM Transactions on Graphics}, vol. 39, no. 6, 2020.

\bibitem{song2021pareidolia}
Linsen Song, Wayne Wu, Chaoyou Fu, Chen Qian, Chen~Change Loy, and Ran He,
\newblock ``Pareidolia face reenactment,''
\newblock in {\em IEEE Conference on Computer Vision and Pattern Recognition (CVPR)}, 2021.

\bibitem{xu2022motion}
Borun Xu, Biao Wang, Jinhong Deng, Jiale Tao, Tiezheng Ge, Yuning Jiang, Wen Li, and Lixin Duan,
\newblock ``Motion and appearance adaptation for cross-domain motion transfer,'' 2022.

\bibitem{kim2021animeceleb}
Kangyeol Kim, Sunghyun Park, Jaeseong Lee, Sunghyo Chung, Junsoo Lee, and Jaegul Choo,
\newblock ``Animeceleb: Large-scale animation celebheads dataset for head reenactment,''
\newblock in {\em Proc. of the European Conference on Computer Vision (ECCV)}, 2022.

\bibitem{rombach2021highresolution}
Robin Rombach, Andreas Blattmann, Dominik Lorenz, Patrick Esser, and Björn Ommer,
\newblock ``High-resolution image synthesis with latent diffusion models,'' 2021.

\bibitem{liu2022compositional}
Nan Liu, Shuang Li, Yilun Du, Antonio Torralba, and Joshua~B Tenenbaum,
\newblock ``Compositional visual generation with composable diffusion models,''
\newblock in {\em Computer Vision--ECCV 2022: 17th European Conference, Tel Aviv, Israel, October 23--27, 2022, Proceedings, Part XVII}. Springer, 2022, pp. 423--439.

\bibitem{Yaniv_2019_Face}
Jordan Yaniv, Yael Newman, and Ariel Shamir,
\newblock ``The face of art: Landmark detection and geometric style in portraits,''
\newblock {\em ACM Transactions on Graphics (TOG)}, vol. 38, no. 4, pp. 1--15, 2019.

\bibitem{Onizuka_2019_ICCV}
Hayato Onizuka, Diego Thomas, Hideaki Uchiyama, and Rin-ichiro Taniguchi,
\newblock ``Landmark-guided deformation transfer of template facial expressions for automatic generation of avatar blendshapes,''
\newblock in {\em Proceedings of the IEEE/CVF International Conference on Computer Vision (ICCV) Workshops}, Oct 2019.

\bibitem{Yuan22NeRFEditing}
Yu-Jie Yuan, Yang-Tian Sun, Yu-Kun Lai, Yuewen Ma, Rongfei Jia, and Lin Gao,
\newblock ``Nerf-editing: Geometry editing of neural radiance fields,''
\newblock in {\em Computer Vision and Pattern Recognition (CVPR)}, 2022.

\bibitem{xu2022deforming}
Tianhan Xu and Tatsuya Harada,
\newblock ``Deforming radiance fields with cages,''
\newblock in {\em ECCV}, 2022.

\bibitem{peng2021CageNeRF}
Yicong Peng, Yichao Yan, Shenqi Liu, Yuhao Cheng, Shanyan Guan, Bowen Pan, Guangtao Zhai, and Xiaokang Yang,
\newblock ``Cagenerf: Cage-based neural radiance fields for genrenlized 3d deformation and animation,''
\newblock in {\em Thirty-Sixth Conference on Neural Information Processing Systems}, 2022.

\bibitem{li2023interactive}
Shaoxu Li and Ye~Pan,
\newblock ``Interactive geometry editing of neural radiance fields,'' 2023.

\bibitem{Siarohin2019FOMM}
Aliaksandr Siarohin, Stéphane Lathuilière, Sergey Tulyakov, Elisa Ricci, and Nicu Sebe,
\newblock ``First order motion model for image animation,''
\newblock in {\em Conference on Neural Information Processing Systems (NeurIPS)}, December 2019.

\bibitem{hong2023dagan}
Fa-Ting Hong, , Li~Shen, and Dan Xu,
\newblock ``Dagan++: Depth-aware generative adversarial network for talking head video generation,''
\newblock {\em IEEE Transactions on Pattern Analysis and Machine Intelligence (TPAMI)}, 2023.

\bibitem{wang2022latent}
Yaohui Wang, Di~Yang, Francois Bremond, and Antitza Dantcheva,
\newblock ``Latent image animator: Learning to animate images via latent space navigation,''
\newblock in {\em International Conference on Learning Representations}, 2022.

\bibitem{Heusel2017GANs}
Martin Heusel, Hubert Ramsauer, Thomas Unterthiner, Bernhard Nessler, and Sepp Hochreiter,
\newblock ``Gans trained by a two time-scale update rule converge to a local nash equilibrium,''
\newblock in {\em Proceedings of the 31st International Conference on Neural Information Processing Systems}, Red Hook, NY, USA, 2017, NIPS'17, p. 6629–6640, Curran Associates Inc.

\bibitem{Narvekar2011CPBD}
Niranjan~D. Narvekar and Lina~J. Karam,
\newblock ``A no-reference image blur metric based on the cumulative probability of blur detection (cpbd),''
\newblock {\em IEEE Transactions on Image Processing}, vol. 20, no. 9, pp. 2678--2683, 2011.

\end{thebibliography}

\end{document}